\documentclass[10pt,twocolumn,letterpaper]{article}

\usepackage[pagenumbers]{cvpr} 
\usepackage[table]{xcolor}
\usepackage{booktabs}    
\usepackage{multirow}    
\usepackage{makecell}    
\usepackage{array}
\usepackage{xcolor,colortbl,pifont}
\usepackage{graphicx}

\definecolor{cvprblue}{rgb}{0.21,0.49,0.74}
\usepackage[pagebackref,breaklinks,colorlinks,allcolors=cvprblue]{hyperref}

\def\paperID{21504} 
\def\confName{CVPR}
\def\confYear{2026}

\title{RegionRoute: Regional Style Transfer with Diffusion Model}

\author{
Bowen Chen$^{1}$\quad
Jake Zuena$^{2}$\quad
Alan C. Bovik$^{1}$\quad
Divya Kothandaraman$^{2}$\\[4pt]
$^{1}$The University of Texas at Austin\quad
$^{2}$Dolby Laboratories, Inc.\\[4pt]
\footnotesize
\texttt{https://github.com/bwchen05/RegionRoute}
}

\begin{document}
\twocolumn[{
\renewcommand\twocolumn[1][]{#1}
\maketitle
\begin{center}
    \centering
    \includegraphics[width=\textwidth]{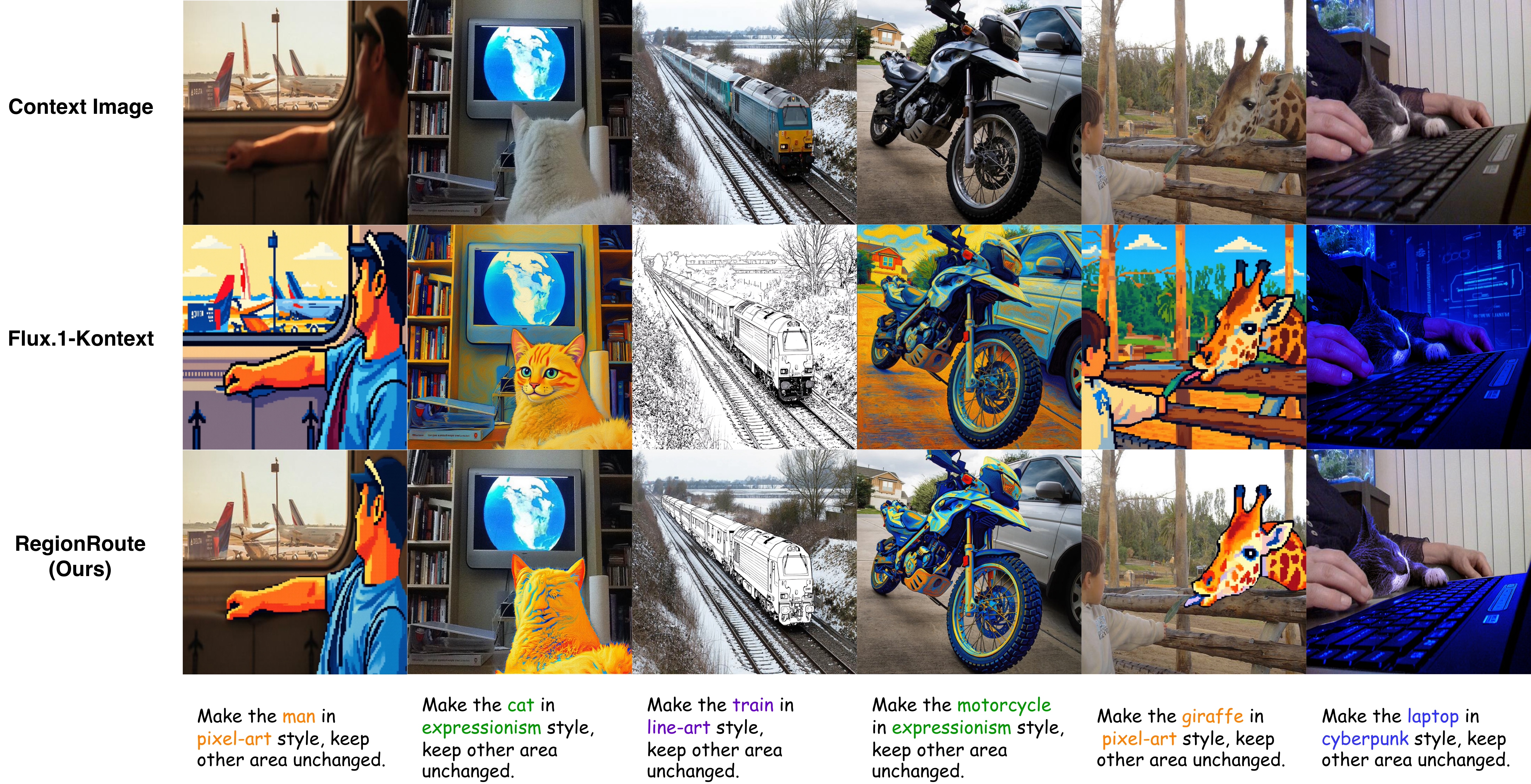}
    
    \captionsetup[figure]{hypcap=false}
    \captionof{figure}{When provided with region-specific editing instructions, our \textbf{RegionRoute} framework more precisely interprets localized style modification prompts and produces visually coherent results. Given prompts such as “Make the man in pixel-art style and keep other areas unchanged,” the baseline image editing model tends to either apply the style globally or distort unrelated regions. Each column shows, from top to bottom: the input context image, the baseline, Flux.1-Kontext~\cite{flux-kontext} output, and our RegionRoute output.}
\label{fig:examples}
\end{center}
}]

\begin{abstract}

\vspace{-8pt}
Precise spatial control in diffusion-based style transfer remains challenging. This challenge arises because diffusion models treat style as a global feature and lack explicit spatial grounding of style representations, making it difficult to restrict style application to specific objects or regions. To our knowledge, existing diffusion models are unable to perform true localized style transfer, typically relying on handcrafted masks or multi-stage post-processing that introduce boundary artifacts and limit generalization. To address this, we propose an attention-supervised diffusion framework that explicitly teaches the model where to apply a given style by aligning the attention scores of style tokens with object masks during training. Two complementary objectives, a Focus loss based on KL divergence and a Cover loss using binary cross-entropy, jointly encourage accurate localization and dense coverage. A modular LoRA-MoE design further enables efficient and scalable multi-style adaptation. To evaluate localized stylization, we introduce the Regional Style Editing Score, which measures Regional Style Matching through CLIP-based similarity within the target region and Identity Preservation via masked LPIPS and pixel-level consistency on unedited areas. Experiments show that our method achieves mask-free, single-object style transfer at inference, producing regionally accurate and visually coherent results that outperform existing diffusion-based editing approaches.

\end{abstract}    

\section{Introduction}
\label{sec:intro}

\begin{figure*}[t]
    \centering
    \includegraphics[width=0.93\linewidth]{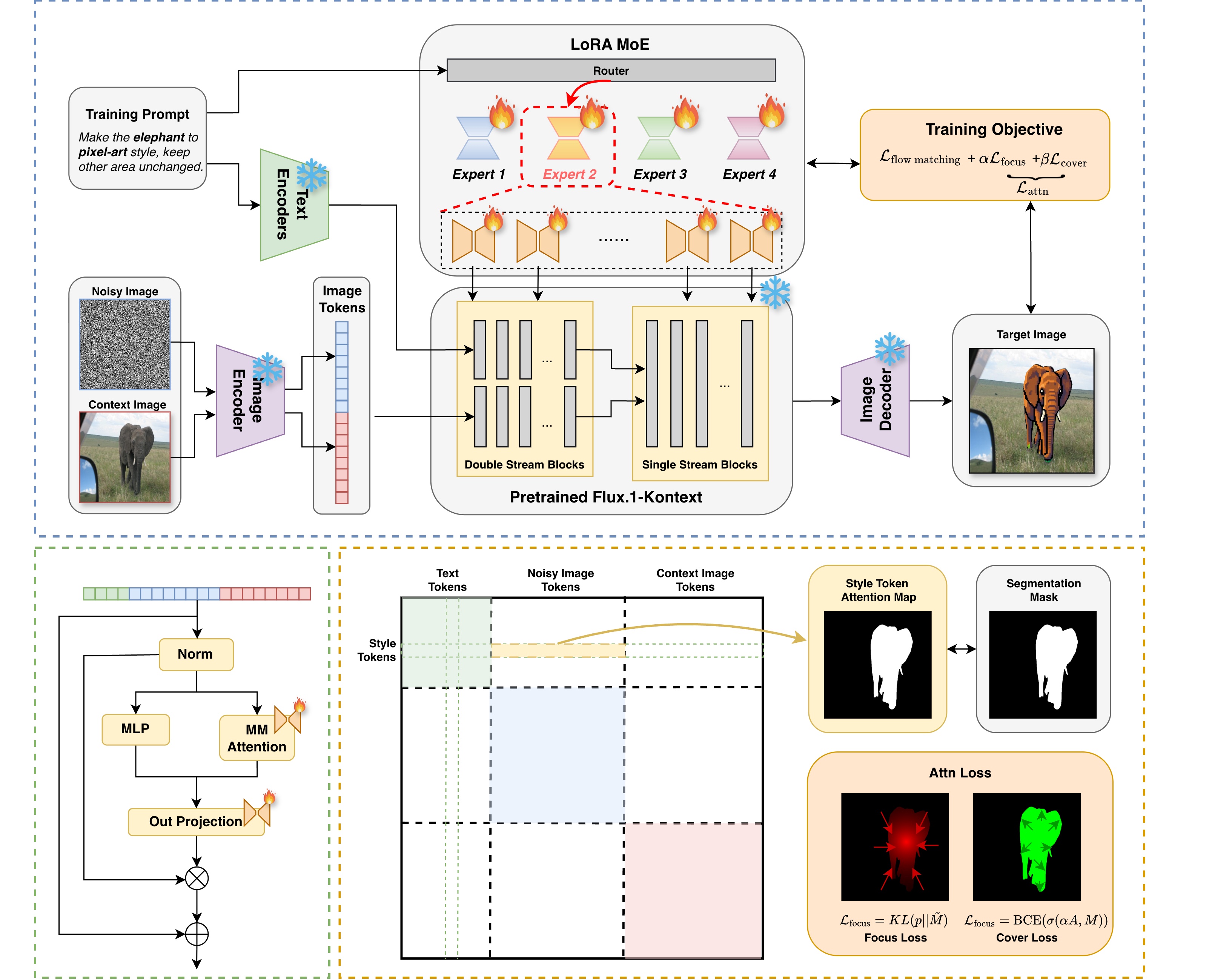}
    \caption{\textbf{Overview of the proposed framework.}
    The upper figure illustrates the overall pipeline based on the pretrained Flux.1-Kontext~\cite{flux-kontext}. Given a context image, a noisy input, and a regional style prompt, image and text tokens are processed through Flux.1-Kontext, where LoRA-MoE modules~\cite{loramoe} adapt attention and projection layers for style-specific learning. The model is optimized with flow matching, focus loss, and cover loss to reconstruct the target stylized image. Style-related attention maps are guided by binary masks, where focus loss concentrates attention within the target area and cover loss ensures spatial coverage for precise localized stylization.}
    \label{fig:pipeline}
\end{figure*}

Recent advances in diffusion-based generative models have shown remarkable capabilities in producing high-quality and diverse visual content across various domains~\cite{ho2020denoising, nichol2021improved, dhariwal2021diffusion, song2020score, saharia2022photorealistic, ramesh2022hierarchical, sd, sdxl, sd3, flux}. 
Models from the Stable Diffusion~\cite{sd, sdxl, sd3} and Flux~\cite{flux, flux-kontext} series achieve strong perceptual realism and semantic consistency, supporting a wide range of tasks such as image generation, editing, and style transfer. Building upon these foundations, some diffusion-based editing approaches~\cite{prompt2prompt, instructpix2pix, masactrl, plugandplay} have further enabled controllable modification of visual attributes guided by text or reference images. Similarly, diffusion-driven style transfer methods~\cite{stylediffusion, diffusionclip, styledrop, StyleID, stylestudio, d2styler} demonstrate strong ability in transferring artistic styles and visual aesthetics.

Despite these advances, precise spatial control over where a style is applied remains an open challenge. Existing style transfer models typically treat style as a global feature~\cite{stylegan, stylegan2, adain, stylediffusion, diffusionclip, styledrop, StyleID, stylealigned, stylestudio, d2styler}, applying it uniformly across the entire image without considering spatial boundaries. As a result, they are unable to modify specific objects or regions. Consequently, the only existing way to achieve localized style effects is through a two-stage pipeline: first performing a global style transfer on the entire image, and then using handcrafted masks to splice the stylized regions with the original content. While this strategy can roughly localize style effects, it introduces several limitations, such as the need for precise mask preparation and visible seams at boundaries. These limitations hinder generalization and restrict the practicality of existing methods.

In principle, diffusion models inherently learn attention maps that capture spatial correspondences between textual concepts and image regions~\cite{prompt2prompt, attendandexcite, paintbyexample, controlnet, t2i}. However, such attentions are not explicitly guided to associate style concepts with specific objects. Therefore, even though the model “sees” the correct regions, it often fails to apply the style precisely, leading to global style shifts.

To address these limitations, we propose a novel attention-supervised diffusion framework that explicitly teaches the model where to apply a style by binding the attention maps of style tokens to the binary masks of target objects during training. Specifically, we fine-tune the pre-trained Flux.1-Kontext~\cite{flux-kontext} backbone using a parameter-efficient LoRA-MoE (Mixture-of-Experts LoRA)~\cite{loramoe} strategy, and enforce KL-divergence and binary cross-entropy losses to match attention distributions with corresponding object masks. This supervision directly establishes a correspondence between style tokens and semantic regions, enabling the model to internalize localized style grounding. As a result, it achieves mask-free, single-object style editing at inference without requiring explicit segmentation or external spatial controls. Some results are shown in Figure~\ref{fig:examples}.

Moreover, existing evaluation protocols primarily measure global style similarity or overall perceptual quality, which do not reflect how well a model performs localized style transfer. To fill this gap, we design a new evaluation metric that quantifies localized style fidelity and unedited region preservation, providing a more comprehensive and objective assessment of spatially controllable style transfer. 

In summary, our contributions are:
\begin{itemize} 
    \item We propose an attention-guided training paradigm that explicitly aligns style token attentions with object masks, enabling precise and mask-free localized style transfer. 
    \item We introduce a LoRA-MoE strategy for parameter-efficient fine-tuning, allowing multiple experts to specialize in diverse styles while keeping the model lightweight and stable. 
    \item We design a new evaluation metric to quantitatively measure localized style fidelity and unedited region preservation, filling a key gap in current evaluation standards. 
\end{itemize}
\section{Related Work}
\label{sec:related_work}

\paragraph{Diffusion-based Image Editing.}
Diffusion-based image editing has rapidly advanced by leveraging the strong generative priors of pretrained diffusion models. Early approaches such as SDEdit~\cite{meng2021sdedit}, Blended Diffusion~\cite{avrahami2022blended}, and Stable Diffusion Inpainting~\cite{sd} perform guided denoising or masked inpainting to enable localized edits while preserving global structure. Later methods including ControlNet~\cite{controlnet}, T2I-Adapter~\cite{t2i}, and BrushNet~\cite{brushnet} incorporate structural priors (e.g., edges, depth, segmentation) for precise spatial control, but rely on external supervision such as masks or sketches. Another line of work achieves implicit localization via prompt manipulation or attention modulation. Prompt-to-Prompt~\cite{prompt2prompt} and InstructPix2Pix~\cite{instructpix2pix} enable text-driven editing, while AnyEdit~\cite{anyedit}, DiffEditor~\cite{diffeditor}, ICEdit~\cite{icedit}, and MGIE~\cite{mgie} generalize this paradigm to multimodal or instruction-based editing. Recent large frameworks such as Flux.1-Kontext~\cite{flux-kontext} and Qwen-Image-Edit~\cite{qwenimageedit} unify these ideas through hierarchical attention and multimodal conditioning, yet still depend on cross-attention to localize edits. Attend-and-Excite~\cite{attendandexcite} modifies attention activations to strengthen underrepresented regions while TokenCompose~\cite{tokencompose} supervises cross-attention maps to bind textual tokens with visual objects. Our work follows this direction by leveraging attention scheme within the diffusion backbone to achieve region-aware style transfer.
\vspace*{-2ex}
\paragraph{Diffusion-based Style Transfer.} Classical neural style transfer (NST) methods, beginning with Gatys \etal~\cite{gatys2016image}, optimize global content and style statistics extracted from CNN features. Feed-forward variants such as AdaIN~\cite{adain}, WCT~\cite{li2017wct}, and SANet~\cite{park2019sanet}, and transformer-based StyTR$^{2}$~\cite{deng2022stytr2}, accelerate stylization by matching global feature statistics. Diffusion-based approaches leverage strong generative priors for high-fidelity stylization. InST~\cite{zhang2023inversion}, StyleDiffusion~\cite{stylediffusion}, and FreeStyle~\cite{he2024freestyle} guide pretrained diffusion models using textual or latent style embeddings, while D$^{2}$Styler~\cite{d2styler} improve stability or attempt localized control via attention injection. Style Injection in Diffusion~\cite{chung2024style} performs training-free key–value replacement to transfer global style. STAM~\cite{fahim2025stam} employs attention modulation for zero-shot transfer, U-StyDiT~\cite{zhang2025u} utilizes diffusion transformers for high-resolution stylization, and StyleStudio~\cite{stylestudio} enables selective manipulation of stylistic elements. While promising, these methods model style as a global latent feature, offering limited region-level controllability. In contrast, our framework performs region-aware style transfer by aligning attention maps of style tokens and object binary mask, enabling localized style modulation without prompts, masks, or external supervision.
\section{Method}
\label{sec:method}

\subsection{Overview}

Our goal is to enable a diffusion model to automatically determine where  a visual style should be applied within an image, e.g., applying a pixel-art style only to a cat without explicit segmentation masks during inference. We achieve this through attention-guided training, which supervises the model’s internal attention maps during fine-tuning to establish spatial correspondence between style tokens and object regions.

Built upon Flux.1-Kontext~\cite{flux-kontext}, a DiT-based diffusion model with joint text–image self-attention, our framework introduces attention supervision that aligns the attention maps of style tokens with target object masks. This correspondence is enforced via dedicated loss terms, allowing the model to ground style concepts to visual regions and perform mask-free localized style transfer at inference.

To efficiently handle multiple styles, we integrate a LoRA-MoE mechanism~\cite{loramoe}, where each style is represented by a specialized LoRA expert attached to a shared diffusion backbone. The backbone learns \emph{where} to apply the style, while each expert defines \emph{how} the style is rendered, enabling modular, plug-and-play control without retraining or cross-style interference. The overview of the proposed framework are shown in Figure~\ref{fig:pipeline}.

\subsection{Attention Map Extraction}

Flux.1-Kontext~\cite{flux-kontext} employs a transformer-based diffusion backbone, where each DiT block performs multi-head self-attention jointly over both image and text tokens. Given a text prompt containing a style phrase (e.g., ``pixel-art style''), we extract the \emph{text-to-image attention slice} associated with the style token \(s\),  which measures how each image token attends to the textual concept representing the target style (Figure~\ref{fig:pipeline}).

For each attention layer \(\ell\) with multi-head attention \(A^{(\ell)} \in \mathbb{R}^{H\times N\times N}\), we first isolate the attention from the image queries \(Q_{\text{img}}\) to the style tokens \(K_s\). We then average over heads, layers, and style tokens to obtain the aggregated style-conditioned attention map:

\begin{equation}
\hat{M}_s = 
\frac{1}{L}\sum_{\ell\in\mathcal{L}}
\frac{1}{H}\sum_{h=1}^{H}
\frac{1}{|K_s|}\sum_{k\in K_s}
A^{(\ell)}_{h}[Q_{\text{img}},k],
\end{equation}
where \(L\) is the set of layers used for supervision, and \(H\) is the number of attention heads. The resulting map \(\hat{M}_s \in \mathbb{R}^{h\times w}\) captures how strongly each spatial token attends to the style token \(s\).
For supervision, the ground-truth mask \(M_s \in [0,1]^{h\times w}\) is obtained by downsampling the object segmentation map to match the attention map.

\subsection{Attention Supervision Losses}

To teach the model to attend accurately and comprehensively to the style-relevant region, we introduce two complementary losses:

\begin{itemize}
    \item \textbf{Focus Loss} — aligns the overall spatial distribution of attention mass with the target object.
    \item \textbf{Cover Loss} — enforces uniform coverage within the object region, discouraging sparse or partial attention.
\end{itemize}

\vspace*{-2ex}
\paragraph{Focus Loss.}
The focus loss aligns the global shape of the predicted attention with the ground-truth mask. We interpret both the predicted attention and mask as normalized probability distributions and minimize their Kullback–Leibler divergence:

\begin{equation}
\label{eq:focus}
\mathcal{L}_{\mathrm{focus}}
=\sum_{s=1}^S
\mathrm{KL}\!\Big(
\mathrm{softmax}(\hat{M}_s/\tau)
\;\Big\|\;
\mathrm{norm}(M_s)
\Big),
\end{equation}
where \(\mathrm{norm}(Z)=\frac{Z}{\sum Z}\) and \(\tau\) controls the sharpness of the attention distribution. This encourages the attention map of each style token to concentrate its mass in the same spatial region as the corresponding object.

\vspace*{-2ex}
\paragraph{Cover Loss.}
While the focus loss ensures global alignment, it does not prevent the model from collapsing attention to a small part of the object. To encourage spatially dense and uniform coverage, we introduce a binary cross-entropy loss that operates at the token level:

\begin{equation}
\label{eq:cover}
\mathcal{L}_{\mathrm{cover}}
=\sum_{s=1}^S
\mathrm{BCE\_logits}\!\big(\alpha\,\hat{M}_s,\ M_s\big),
\end{equation}
where \(\mathrm{BCE\_logits}\) is the numerically stable binary cross-entropy operating on logits, and \(\alpha\) is a contrast factor that amplifies attention magnitude for stronger gradients.  
This term penalizes attention outside the object region (\(M_s = 0\)) and rewards attention inside (\(M_s = 1\)), producing smooth and coherent attention over the object.

Together, these two objectives ensure that the model learns not only to localize the correct style-relevant region (via $\mathcal{L}_{\mathrm{focus}}$) but also to distribute attention densely within it (via $\mathcal{L}_{\mathrm{cover}}$), 
resulting in coherent and spatially consistent style application.

\subsection{LoRA-MoE Adaptation}
To efficiently support multiple visual styles, we introduce a modular Low-Rank Adaptation with LoRA-MoE~\cite{loramoe} scheme. Instead of fine-tuning a single LoRA across all styles, which often causes interference and degraded style fidelity, we assign each style a lightweight LoRA expert trained independently on the same shared diffusion backbone.  During training, only the expert corresponding to the current style is activated, while the backbone remains frozen to preserve the attention-grounded spatial reasoning learned earlier. At inference, the appropriate expert is selected based on the target style token, enabling plug-and-play style control.  
This design provides three key benefits:
(i) parameter efficiency — new styles are added without retraining the backbone,
(ii) specialization — each expert learns distinct style patterns,
and (iii) stability — the shared backbone ensures consistent spatial alignment across experts.

\subsection{Training Objective}
The overall objective combines the diffusion reconstruction loss with attention supervision:
\begin{equation}
\mathcal{L}
=\mathcal{L}_{\epsilon}
+\lambda_{f}\mathcal{L}_{\mathrm{focus}}
+\lambda_{c}\mathcal{L}_{\mathrm{cover}},
\end{equation}
where $\mathcal{L}_{\epsilon}=\|\hat{\epsilon}-\epsilon\|_2^2$ is the standard noise prediction loss, and $\lambda_{f}, \lambda_{c}$ balance the relative strengths of attention alignment and coverage.
\section{Regional Style Editing Score}
\label{metric}

Existing metrics such as FID~\cite{fid} or CLIP-similarity~\cite{clip} mainly capture global appearance, lacking sensitivity to whether the style is accurately localized and whether unedited regions are preserved. To address this gap, we introduce the Regional Style Editing Score (RSE-Score), a metric specifically designed for evaluating \emph{single-object regional style transfer}. RSE-Score decomposes the evaluation into two complementary aspects: (1) \emph{Regional Style Matching (RSM)}, assessing whether the target region successfully reflects the intended style, and (2) \emph{Identity Preservation} components, evaluating the perceptual and pixel-level fidelity of unedited areas. Together, these measures provide a comprehensive and interpretable assessment of both local style accuracy and spatial controllability.

Let $x$ denote the original image, $\hat{x}$ the edited image, and $M \in \{0,1\}^{H \times W}$ the binary target mask, where $M_p=1$ if pixel $p$ belongs to the object region and $M_p=0$ otherwise. The complementary background region is denoted by $(1 - M)$.

\subsection{Regional Style Matching (RSM)}
RSM quantifies how well the style within the target region matches the desired textual description. Instead of relying on patch-level features, we crop the edited image to the minimal bounding box enclosing the target mask (with a small padding margin) and compute its similarity to the style text using CLIP~\cite{clip} encoders:
\begin{equation}
\text{RSM}
=
\frac{1}{2}
\big(
1
+
\cos\!\big(
f_{\text{img}}(\hat{x}_{\text{crop}}),
f_{\text{text}}(s)
\big)
\big),
\label{eq:rsm}
\end{equation}
where $\hat{x}_{\text{crop}}$ is the cropped region around the object, and $f_{\text{img}}(\cdot)$, $f_{\text{text}}(\cdot)$ denote CLIP’s image and text feature extractors. The cosine similarity is linearly mapped to $[0,1]$ for interpretability. This formulation focuses the style evaluation strictly within the edited region, mitigating interference from unrelated background areas.

\subsection{Identity Preservation}
To separately assess fidelity in unedited regions, we compute two complementary metrics on the background:

\vspace*{-2ex}
\paragraph{Perceptual Consistency (LPIPS).}
We use the spatial version of LPIPS~\cite{lpips} to measure perceptual distance between the edited and original images within the unedited region:
\begin{equation}
\text{LPIPS}_{\text{bg}}
=
\frac{
\sum_{p}
(1 - M_p)\,
\text{LPIPS}_p(\hat{x}, x)
}{
\sum_{p}
(1 - M_p)
},
\label{eq:lpips}
\end{equation}
where $\text{LPIPS}_p$ denotes the per-pixel perceptual difference. Lower values indicate better background preservation.
\vspace*{-2ex}
\paragraph{Pixel Consistency (MSE).}
In addition, we compute a masked mean squared error over the same background region:
\begin{equation}
\text{MSE}_{\text{bg}}
=
\frac{
\sum_{p}
(1 - M_p)\,
\|\hat{x}_p - x_p\|_2^2
}{
\sum_{p}
(1 - M_p)
}.
\label{eq:mse}
\end{equation}
This term captures fine-grained pixel alignment and complements the perceptual similarity measure.

Unlike the previous unified identity score, we now report $\text{LPIPS}_{\text{bg}}$ and $\text{MSE}_{\text{bg}}$ independently, offering a clearer diagnostic view of both perceptual and structural preservation. In summary, RSM evaluates style correctness within the edited region, while LPIPS\(_{\text{bg}}\) and MSE\(_{\text{bg}}\) quantify background fidelity—together forming a comprehensive benchmark for regional style transfer quality.

\begin{figure*}[t]
    \centering
    
    \includegraphics[width=\linewidth]{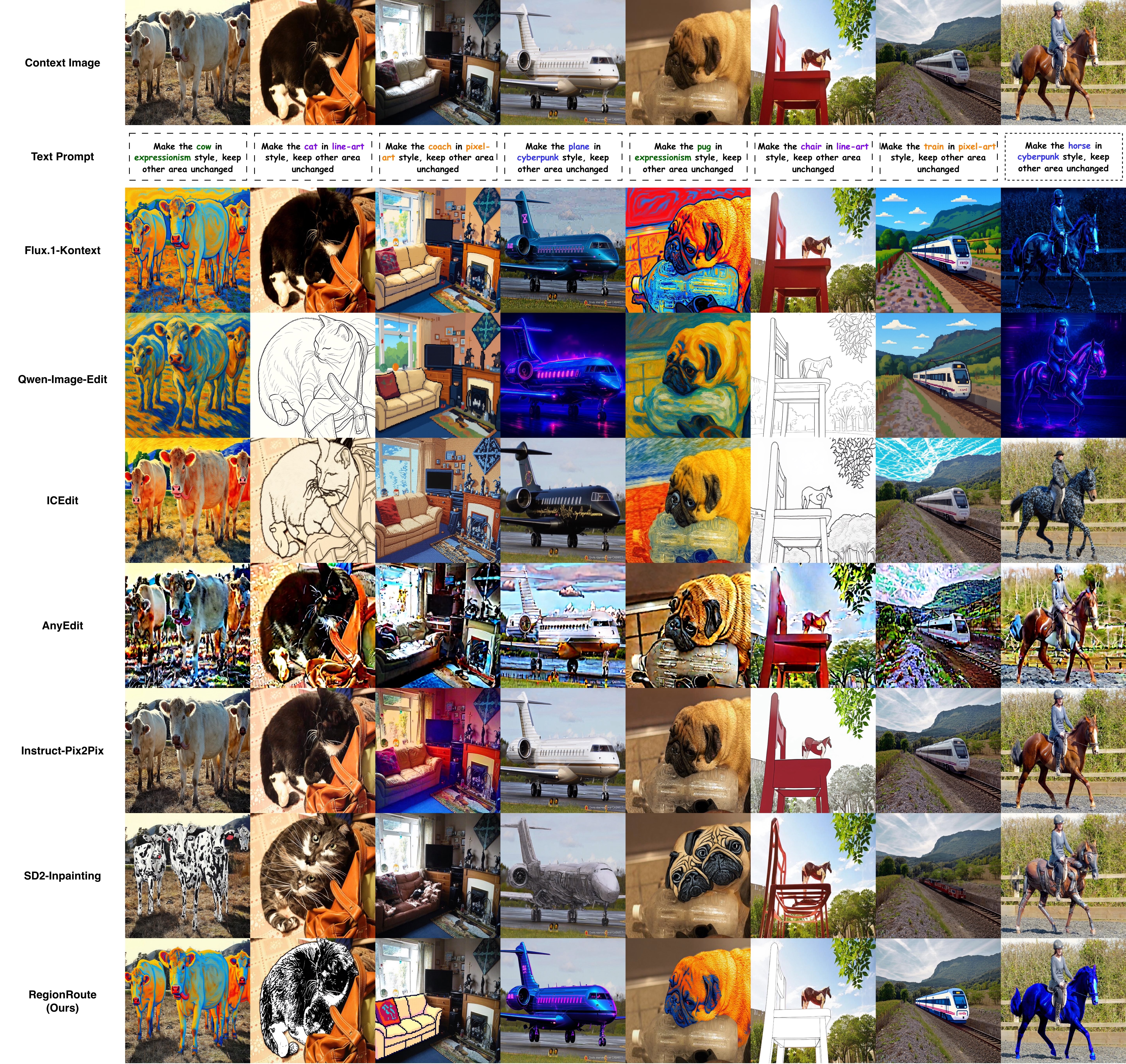}
    \caption{Qualitative comparison of state-of-the-art instruction-based image editing methods.}
    \label{fig:qualitative_results}
\end{figure*}
\section{Experiments}
\label{sec:experiments}

\definecolor{rank1}{gray}{0.85}
\definecolor{rank2}{gray}{0.90}
\definecolor{rank3}{gray}{0.95}

\begin{table*}[t]
    \centering
    \caption{Quantitative comparison with baseline methods across three datasets. We report mean\textsubscript{std} for RSM (↑), LPIPS\textsubscript{bg} (↓), and MSE\textsubscript{bg} (↓). Darker cells indicate better ranks (1–3) within each dataset and metric.}
    \label{tab:quantitative_results}
    \resizebox{\textwidth}{!}{
    \begin{tabular}{l|ccc|ccc|ccc}
        \toprule
        \multirow{2}{*}{\textbf{Model}} 
        & \multicolumn{3}{c|}{\textbf{COCO}} 
        & \multicolumn{3}{c|}{\textbf{Pascal VOC}} 
        & \multicolumn{3}{c}{\textbf{BIG}} \\
        \cmidrule(lr){2-4} \cmidrule(lr){5-7} \cmidrule(lr){8-10}
        & \textbf{RSM} $\uparrow$ & \textbf{$\text{LPIPS}_{\text{bg}}$} $\downarrow$ & \textbf{$\text{MSE}_{\text{bg}}$} $\downarrow$ 
        & \textbf{RSM} $\uparrow$ & \textbf{$\text{LPIPS}_{\text{bg}}$} $\downarrow$ & \textbf{$\text{MSE}_{\text{bg}}$} $\downarrow$
        & \textbf{RSM} $\uparrow$ & \textbf{$\text{LPIPS}_{\text{bg}}$} $\downarrow$ & \textbf{$\text{MSE}_{\text{bg}}$} $\downarrow$ \\
        \midrule
        Flux.1-Kontext~\cite{flux-kontext} 
        & \cellcolor{rank3}0.6126\textsubscript{0.0180} & 0.4546\textsubscript{0.2960} & 0.1699\textsubscript{0.2254} 
        & \cellcolor{rank3}0.6135\textsubscript{0.0170} & 0.4326\textsubscript{0.3135} & 0.1455\textsubscript{0.2258} 
        & \cellcolor{rank2}0.6169\textsubscript{0.0228} & 0.4815\textsubscript{0.3125} & 0.1981\textsubscript{0.2408} \\

        Qwen-Image-Edit~\cite{qwenimageedit} 
        & \cellcolor{rank1}0.6235\textsubscript{0.0164} & 0.7530\textsubscript{0.1605} & 0.4398\textsubscript{0.3876} 
        & \cellcolor{rank1}0.6333\textsubscript{0.0104} & 0.8078\textsubscript{0.1550} & 0.4477\textsubscript{0.3767} 
        & \cellcolor{rank1}0.6329\textsubscript{0.0103} & 0.7185\textsubscript{0.1687} & 0.4284\textsubscript{0.3731} \\

        ICEdit~\cite{icedit} 
        & 0.6086\textsubscript{0.0152} & 0.3512\textsubscript{0.1979} & 0.1568\textsubscript{0.2800} 
        & 0.6102\textsubscript{0.0147} & 0.3506\textsubscript{0.1963} & 0.1284\textsubscript{0.2304} 
        & 0.6115\textsubscript{0.0164} & 0.4282\textsubscript{0.2315} & 0.1570\textsubscript{0.2697} \\

        AnyEdit~\cite{anyedit} 
        & 0.6085\textsubscript{0.0104} & 0.6895\textsubscript{0.1564} & 0.2633\textsubscript{0.2913} 
        & 0.6081\textsubscript{0.0107} & 0.6834\textsubscript{0.1715} & 0.2345\textsubscript{0.2454} 
        & 0.6078\textsubscript{0.0097} & 0.7109\textsubscript{0.1679} & 0.2446\textsubscript{0.2813} \\

        Instruct-Pix2Pix~\cite{instructpix2pix} 
        & 0.5978\textsubscript{0.0115} & \cellcolor{rank2}0.1867\textsubscript{0.1995} & \cellcolor{rank3}0.0516\textsubscript{0.1587} 
        & 0.5969\textsubscript{0.0111} & \cellcolor{rank3}0.1554\textsubscript{0.1472} & 0.0299\textsubscript{0.1078} 
        & 0.5939\textsubscript{0.0111} & \cellcolor{rank2}0.1466\textsubscript{0.1377} & \cellcolor{rank3}0.0347\textsubscript{0.1403} \\

        SD2-Inpainting~\cite{sd} 
        & 0.6028\textsubscript{0.0135} & \cellcolor{rank1}0.0859\textsubscript{0.0566} & \cellcolor{rank1}0.0039\textsubscript{0.0047} 
        & 0.6094\textsubscript{0.0137} & \cellcolor{rank1}0.0960\textsubscript{0.0410} & \cellcolor{rank1}0.0043\textsubscript{0.0049} 
        & 0.6016\textsubscript{0.0139} & \cellcolor{rank1}0.1019\textsubscript{0.0492} & \cellcolor{rank1}0.0063\textsubscript{0.0073} \\

        \textbf{RegionRoute (Ours)} 
        & \cellcolor{rank2}\textbf{0.6128}\textsubscript{0.0162} & \cellcolor{rank3}\textbf{0.2103}\textsubscript{0.1933} & \textbf{0.0729}\textsubscript{0.1281} 
        & \cellcolor{rank2}\textbf{0.6147}\textsubscript{0.0151} & \cellcolor{rank2}\textbf{0.1331}\textsubscript{0.1429} & \cellcolor{rank3}\textbf{0.0269}\textsubscript{0.0571} 
        & \cellcolor{rank3}\textbf{0.6159}\textsubscript{0.0193} & \cellcolor{rank3}\textbf{0.1593}\textsubscript{0.1681} & \textbf{0.0474}\textsubscript{0.0967} \\
        \bottomrule
    \end{tabular}
    }
\end{table*}

\subsection{Dataset and Pseudo-GT Generation}
\label{subsec:dataset}

We use a subset of the Grounded COCO dataset introduced in TokenCompose~\cite{tokencompose}, built upon MS-COCO~\cite{coco} image–caption pairs. We randomly sample 150 image–caption pairs for fine-tuning and analysis. For each image, one target object (with its binary mask) is selected, and a pseudo ground-truth (pseudo-GT) image is generated by applying a diffusion-based style transfer model and compositing it with the original image. 

Since no existing dataset provides supervision for localized style transfer, pseudo-GT generation serves as a practical strategy to enable spatially aware style learning. Benefiting from Flux.1-Kontext~\cite{flux-kontext}’s strong instance recognition, precise mask alignment is not required, as the model can naturally learn smooth region boundaries. To ensure stylistic diversity, we generate pseudo-GT images in four representative styles—pixel art, cyberpunk, expressionism, and line art—resulting in 600 training samples (150 per style) for fine-tuning. More details of the pseudo-GT generation are explained in Supplementary Materials.

\subsection{Experimental Setup}
\label{subsec:setup}

We fine-tuned the Flux.1-Kontext~\cite{flux-kontext} model using LoRA-MoE~\cite{loramoe} on a single NVIDIA GH200 GPU (120 GB VRAM). Training is performed at $1024\times1024$ resolution using \texttt{bf16} mixed precision and 8-bit Adam optimizer. The LoRA rank is set to 4, with a learning rate of $1\times10^{-4}$, batch size of 2, and gradient accumulation of 4. We train for 5000 steps under a constant learning rate schedule without warmup. Focus and cover losses are weighted by 0.1 and 0.2, respectively.

\subsection{Experiment Results}
\label{subsec:results}

\paragraph{Baselines.} We compare our model with six representative instruction-based image editing approaches spanning different paradigms, including diffusion-based, instruction-following, and vision–language models. Flux.1-Kontext~\cite{flux-kontext} is a recent text-guided editor supporting soft regional conditioning. Qwen-Image-Edit~\cite{qwenimageedit} is an MLLM-based system that interprets natural language instructions to directly produce edited images. ICEdit~\cite{icedit} is an instruction-driven diffusion model for localized edits aligned with text. AnyEdit~\cite{anyedit} is another diffusion editor for controlled modifications. We also include Instruct-Pix2Pix~\cite{instructpix2pix}, and SD2-Inpainting~\cite{sd}, the Stable Diffusion v2 inpainting variant for mask-based regional editing.

\vspace*{-2ex}
\paragraph{Datasets.} We evaluate our method on three benchmark datasets: COCO, Pascal VOC, and BIG. We intentionally select segmentation-based datasets that provide pixel-level object masks, which are essential for evaluating regional style transfer. The availability of accurate masks allows us to precisely define the edited region and its complement for computing the proposed metrics, ensuring consistent and spatially aligned evaluation across all methods. For the COCO dataset, we use the Grounded COCO dataset introduced in~\cite{tokencompose} while excluding all images used for training in our setup; only the remaining images are employed for evaluation. The Pascal VOC dataset~\cite{pascal} follows the re-labeled version introduced in~\cite{cascadepsp}, which provides more accurate and consistent annotations than the original Pascal VOC labels. For the BIG dataset~\cite{cascadepsp}, we use both the official evaluation and test subsets for quantitative and qualitative comparisons.

\vspace*{-2ex}
\paragraph{Results and Analysis}

\definecolor{goodgreen}{RGB}{24,140,70}
\newcommand{\cmark}{\textcolor{goodgreen}{\ding{51}}} 
\newcommand{\xmark}{\textcolor{gray}{\ding{55}}}       
\renewcommand{\arraystretch}{1.15}

\begin{table}[t]
\centering
\caption{
Controllability and semantic reliability evaluation via Vision-Language Model (VLM). Q1: “Is the \emph{object} in the \emph{target style}?”; Q2: “Is the \emph{background} in the \emph{target style}?; Q3: “Is the \emph{object} in the \emph{negative style}?”; Q4: “Is the \emph{background} in the \emph{negative style}?” Each cell shows the probability (\%) of a “Yes” response to the corresponding binary question.}
\label{tab:yes-probability}
\resizebox{0.47\textwidth}{!}{
\setlength{\tabcolsep}{4pt}
\begin{tabular}{l|cccc|cccc|cccc}
\toprule
\multirow{2}{*}{\textbf{Model}} 
& \multicolumn{4}{c|}{\textbf{COCO}} 
& \multicolumn{4}{c|}{\textbf{Pascal VOC}} 
& \multicolumn{4}{c}{\textbf{BIG}} \\
\cmidrule(lr){2-5} \cmidrule(lr){6-9} \cmidrule(lr){10-13}
& Q1$\uparrow$ & Q2$\downarrow$ & Q3$\downarrow$ & Q4$\downarrow$
& Q1$\uparrow$ & Q2$\downarrow$ & Q3$\downarrow$ & Q4$\downarrow$
& Q1$\uparrow$ & Q2$\downarrow$ & Q3$\downarrow$ & Q4$\downarrow$ \\
\midrule
Flux.1-Kontext~\cite{flux-kontext}     
& 0.63 & 0.44 & \cellcolor{rank3}0.08 & 0.06 
& \cellcolor{rank3}0.65 & 0.43 & 0.08 & 0.05 
& \cellcolor{rank3}0.63 & 0.39 & \cellcolor{rank3}0.11 & 0.10 \\

Qwen-Image-Edit~\cite{qwenimageedit}   
& \cellcolor{rank1}0.98 & 0.86 & \cellcolor{rank1}0.01 & \cellcolor{rank1}0.00 
& \cellcolor{rank1}0.97 & 0.78 & \cellcolor{rank1}0.02 & \cellcolor{rank1}0.01 
& \cellcolor{rank1}0.98 & 0.90 & \cellcolor{rank1}0.05 & \cellcolor{rank2}0.04 \\

ICEdit~\cite{icedit}                   
& \cellcolor{rank3}0.72 & 0.40 & 0.11 & \cellcolor{rank3}0.05 
& 0.50 & 0.32 & \cellcolor{rank2}0.06 & 0.04 
& \cellcolor{rank3}0.63 & 0.36 & 0.15 & \cellcolor{rank3}0.06 \\

AnyEdit~\cite{anyedit}                 
& 0.50 & 0.41 & 0.57 & 0.47 
& 0.46 & 0.28 & 0.47 & 0.28 
& 0.58 & 0.50 & 0.58 & 0.47 \\

Instruct-Pix2Pix~\cite{instructpix2pix} 
& 0.06 & \cellcolor{rank3}0.08 & \cellcolor{rank2}0.03 & 0.03 
& 0.10 & \cellcolor{rank2}0.05 & \cellcolor{rank2}0.06 & \cellcolor{rank3}0.03 
& 0.10 & \cellcolor{rank3}0.09 & \cellcolor{rank2}0.06 & 0.07 \\

SD2-Inpainting~\cite{sd}               
& 0.36 & \cellcolor{rank1}0.01 & 0.15 & \cellcolor{rank2}0.01 
& 0.24 & \cellcolor{rank1}0.04 & \cellcolor{rank3}0.07 & \cellcolor{rank2}0.02 
& 0.54 & \cellcolor{rank1}0.04 & 0.32 & \cellcolor{rank1}0.03 \\

\textbf{RegionRoute (Ours)}            
& \cellcolor{rank2}\textbf{0.73} & \cellcolor{rank2}\textbf{0.07} & \textbf{0.12} & \cellcolor{rank1}\textbf{0.00} 
& \cellcolor{rank2}\textbf{0.74} & \cellcolor{rank3}\textbf{0.09} & \textbf{0.10} & \cellcolor{rank2}\textbf{0.02} 
& \cellcolor{rank2}\textbf{0.76} & \cellcolor{rank2}\textbf{0.07} & \textbf{0.23} & \cellcolor{rank2}\textbf{0.04} \\
\bottomrule
\end{tabular}
}
\end{table}

\begin{table*}[t]
\centering
\caption{
Ablation of key loss components, network streams, and LoRA ranks on three datasets.
We report mean\textsubscript{std} for RSM (↑), LPIPS\textsubscript{bg} (↓), and MSE\textsubscript{bg} (↓).
\cmark~denotes the component is enabled.
}
\label{tab:ablation}
\setlength{\tabcolsep}{4pt}
\small
\resizebox{\textwidth}{!}{
\begin{tabular}{lcccccccc ccc ccc}
\toprule
\multicolumn{1}{c}{\textbf{Variant}} &
\multicolumn{5}{c}{\textbf{Enabled Components / Blocks}} &
\multicolumn{3}{c}{\textbf{COCO}} &
\multicolumn{3}{c}{\textbf{Pascal VOC}} &
\multicolumn{3}{c}{\textbf{BIG}} \\
\cmidrule(lr){2-6}\cmidrule(lr){7-9}\cmidrule(lr){10-12}\cmidrule(lr){13-15}
& $\mathcal{L}_{\text{cover}}$ & $\mathcal{L}_{\text{focus}}$ & \textit{Double} & \textit{Single} & \textit{LoRA} &
\textbf{RSM} $\uparrow$ & $\text{LPIPS}_{\text{bg}}$ $\downarrow$ & $\text{MSE}_{\text{bg}}$ $\downarrow$ &
\textbf{RSM} $\uparrow$ & $\text{LPIPS}_{\text{bg}}$ $\downarrow$ & $\text{MSE}_{\text{bg}}$ $\downarrow$ &
\textbf{RSM} $\uparrow$ & $\text{LPIPS}_{\text{bg}}$ $\downarrow$ & $\text{MSE}_{\text{bg}}$ $\downarrow$ \\
\midrule
\rowcolor{gray!8}
\textbf{Ours} 
& \cmark & \cmark & \cmark & \cmark & 4
& \textbf{0.6128}\textsubscript{0.0162} & \textbf{0.2103}\textsubscript{0.1933} & \textbf{0.0729}\textsubscript{0.1281}
& \textbf{0.6147}\textsubscript{0.0151} & \textbf{0.1331}\textsubscript{0.1429} & \textbf{0.0269}\textsubscript{0.0571}
& \textbf{0.6159}\textsubscript{0.0193} & \textbf{0.1593}\textsubscript{0.1681} & \textbf{0.0474}\textsubscript{0.0681} \\
\midrule
\textbf{w/o $\mathcal{L}_{\text{cover}}$}
& \xmark & \cmark & \cmark & \cmark & 4
& 0.6120\textsubscript{0.0163} & 0.2174\textsubscript{0.1889} & 0.0730\textsubscript{0.1069}
& 0.6142\textsubscript{0.0155} & 0.1383\textsubscript{0.1530} & 0.0359\textsubscript{0.0580}
& 0.6152\textsubscript{0.0193} & 0.1605\textsubscript{0.1680} & 0.0488\textsubscript{0.0681} \\
\textbf{w/o $\mathcal{L}_{\text{focus}}$}
& \cmark & \xmark & \cmark & \cmark & 4
& 0.6127\textsubscript{0.0159} & 0.2132\textsubscript{0.1852} & 0.0740\textsubscript{0.1117}
& 0.6147\textsubscript{0.0153} & 0.1359\textsubscript{0.1393} & 0.0325\textsubscript{0.0497}
& 0.6158\textsubscript{0.0190} & 0.1612\textsubscript{0.1620} & 0.0542\textsubscript{0.0917} \\
\midrule
\textbf{w/o Double}
& \cmark & \cmark & \xmark & \cmark & 4
& 0.6168\textsubscript{0.0168} & 0.4225\textsubscript{0.2838} & 0.1409\textsubscript{0.1831}
& 0.6175\textsubscript{0.0155} & 0.3887\textsubscript{0.3064} & 0.0980\textsubscript{0.1435}
& 0.6195\textsubscript{0.0198} & 0.3843\textsubscript{0.3086} & 0.1275\textsubscript{0.1458} \\
\textbf{w/o Single}
& \cmark & \cmark & \cmark & \xmark & 4
& 0.6190\textsubscript{0.0163} & 0.5203\textsubscript{0.3060} & 0.2284\textsubscript{0.2216}
& 0.6185\textsubscript{0.0158} & 0.4252\textsubscript{0.3303} & 0.1390\textsubscript{0.1884}
& 0.6192\textsubscript{0.0205} & 0.3933\textsubscript{0.3110} & 0.1402\textsubscript{0.1595} \\
\midrule
\textbf{Rank = 8}
& \cmark & \cmark & \cmark & \cmark & 8
& 0.6137\textsubscript{0.0154} & 0.2007\textsubscript{0.1739} & 0.0752\textsubscript{0.1491}
& 0.6158\textsubscript{0.0144} & 0.1289\textsubscript{0.1365} & 0.0305\textsubscript{0.0811}
& 0.6169\textsubscript{0.0183} & 0.1378\textsubscript{0.1280} & 0.0387\textsubscript{0.0726} \\
\textbf{Rank = 16}
& \cmark & \cmark & \cmark & \cmark & 16
& 0.6126\textsubscript{0.0150} & 0.1876\textsubscript{0.1583} & 0.0671\textsubscript{0.1229}
& 0.6158\textsubscript{0.0140} & 0.1182\textsubscript{0.1099} & 0.0212\textsubscript{0.0480}
& 0.6152\textsubscript{0.0179} & 0.1177\textsubscript{0.1059} & 0.0265\textsubscript{0.0491} \\
\bottomrule
\end{tabular}
}
\end{table*}

Table~\ref{tab:quantitative_results} presents quantitative comparisons across three datasets, while Figure~\ref{fig:qualitative_results} illustrates qualitative examples.  RSM reflects regional style accuracy, and LPIPS\(_{\text{bg}}\)/MSE\(_{\text{bg}}\) measure background preservation, where higher RSM and lower LPIPS\(_{\text{bg}}\), MSE\(_{\text{bg}}\) indicate better localized editing performance.

Across all datasets, consistent patterns can be observed. Flux.1-Kontext~\cite{flux-kontext} and Qwen-Image-Edit~\cite{qwenimageedit} achieve high RSM values, showing strong style generation ability, but their elevated background distortion indicates a tendency toward global style transfer. ICEdit~\cite{icedit} and AnyEdit~\cite{anyedit} obtain moderate RSM with less stable regional control, while Instruct-Pix2Pix~\cite{instructpix2pix} and SD2-Inpainting~\cite{sd} preserve unedited regions effectively but provide limited stylization strength. Overall, existing methods tend to emphasize either stylistic fidelity or background consistency, but rarely achieve both simultaneously. In contrast, our proposed \textbf{RegionRoute} attains a favorable balance between regional style fidelity and background preservation. It maintains competitive RSM while substantially lowering LPIPS\(_{\text{bg}}\) and MSE\(_{\text{bg}}\), indicating that edits are well localized and semantically coherent. Qualitative results in Figure~\ref{fig:qualitative_results} further confirm that RegionRoute applies the target style precisely within the intended area, preserves the structure of unedited regions, and maintains visual harmony across the entire image.

To further assess controllability and semantic reliability, Table~\ref{tab:yes-probability} introduces four binary questions evaluated by Qwen2.5-VL-7B-Instruct~\cite{qwen25}:  
(Q1) “Is the \emph{object} in the \emph{target style}?” (higher is better);  
(Q2) “Is the \emph{background} in the \emph{target style}?” (lower is better, indicating less style leakage);  
(Q3) “Is the \emph{object} in the \emph{negative style}?”; and  
(Q4) “Is the \emph{background} in the \emph{negative style}?”  Q3--Q4 serve as sanity checks to ensure the model does not produce false positives or semantically inconsistent outputs. \textbf{RegionRoute} achieves high Q1 probabilities with minimal Q2 leakage and very low Q3--Q4 values across all datasets, reflecting accurate regional stylization, strong semantic reliability, and minimal background contamination. Most baseline methods show consistent behavior, while AnyEdit exhibits elevated Q3--Q4 due to its semantically chaotic outputs, which sometimes confuse the vision-language evaluator. Additional visualizations along with failure case analyses are provided in the Supplementary Material.

\subsection{Ablation Studies}
\label{subsec:ablation}

\begin{figure}[t]
    \centering
    \includegraphics[width=\linewidth]{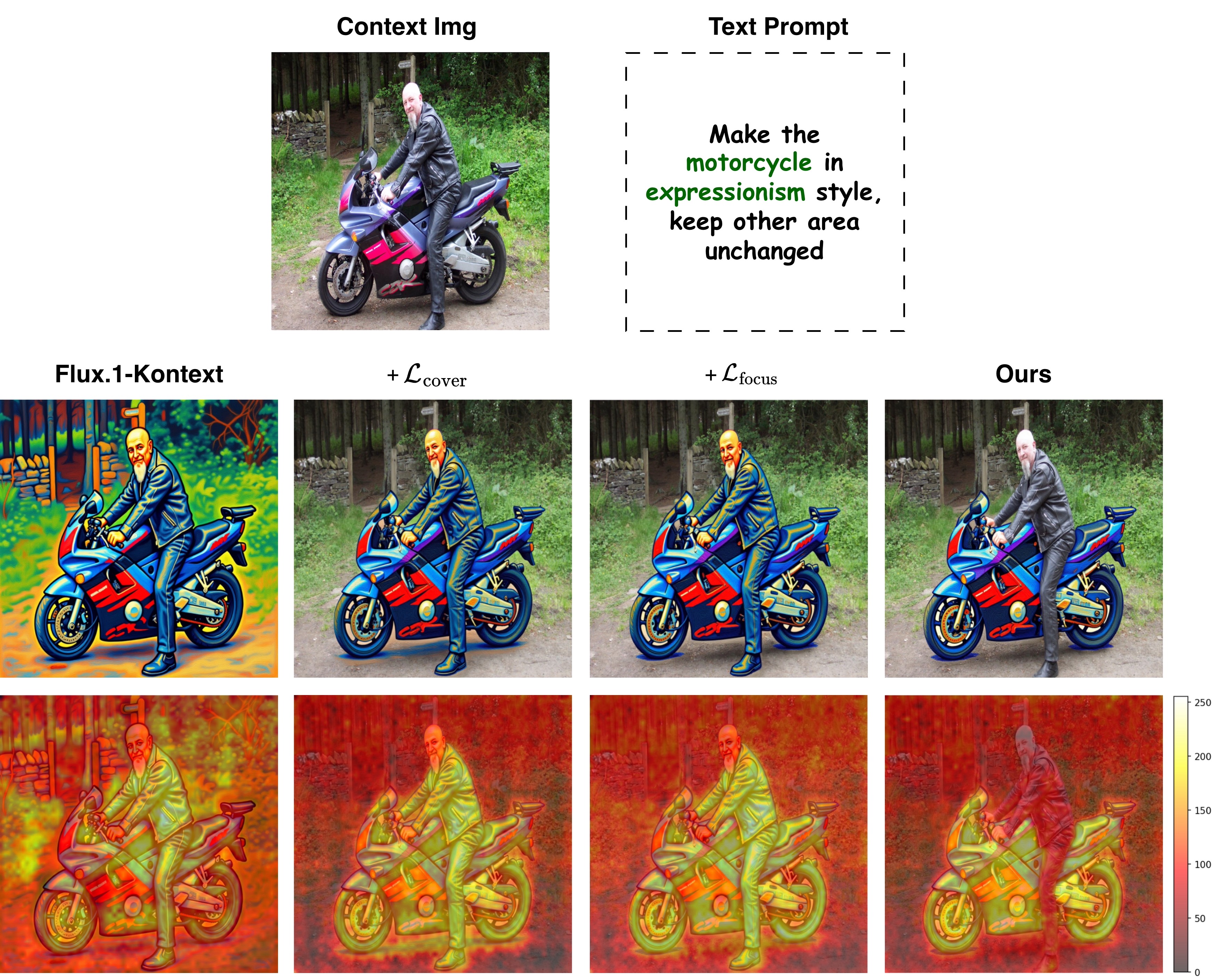}
    \caption{
    Visualization of attention maps under different loss configurations. Using only $\mathcal{L}_{\text{cover}}$ or $\mathcal{L}_{\text{focus}}$ causes attention spillover to nearby area, whereas our full objective focuses on the \textit{motorcycle} without leakage to surrounding regions, demonstrating its ability to maintain precise and consistent attention.
    }
    \label{fig:attn_result}
\end{figure}

To understand the effectiveness of each component in RegionRoute, we conduct a comprehensive ablation study on three datasets. Table~\ref{tab:ablation} summarizes the quantitative results, while Figure~\ref{fig:attn_result} provides qualitative comparisons on each component of our training loss.

Removing either $\mathcal{L}_{\text{cover}}$ or $\mathcal{L}_{\text{focus}}$ leads to a consistent degradation in all metrics across datasets. As shown in Table~\ref{tab:ablation}, both losses contribute to improving regional style matching (RSM$\uparrow$) while reducing background distortion (LPIPS$_{\text{bg}}\downarrow$, MSE$_{\text{bg}}\downarrow$). The $\mathcal{L}_{\text{cover}}$ term encourages the model to preserve context coverage, and $\mathcal{L}_{\text{focus}}$ sharpens object-specific adaptation. Figure~\ref{fig:attn_result} visualizes the attention distributions under different settings. When trained with only $\mathcal{L}_{\text{cover}}$ or $\mathcal{L}_{\text{focus}}$, the attention partially concentrates on the \textit{motorcycle}, but also spills onto nearby humans, leading to inconsistent background reconstruction. In contrast, the full training objective aligns attention exclusively on the \textit{motorcycle} while cleanly suppressing activations around surrounding regions. This confirms that the proposed joint loss formulation effectively enforces precise spatial localization and prevents attention leakage. More visualizations and analysis about the attention loss are in Supplementary Materials.

The Single and Double stream blocks originate from the Flux.1-Kontext~\cite{flux-kontext} architecture, each modeling complementary aspects of spatial context. We investigate whether applying LoRA adaptation~\cite{lora} to these blocks benefits the final performance. As shown in Table~\ref{tab:ablation}, disabling LoRA on either stream (\textit{w/o Double} or \textit{w/o Single}) slightly increases RSM, but substantially worsens LPIPS\textsubscript{bg} and MSE\textsubscript{bg}. This indicates that while the target region may look more “on-style,” the model loses control over the consistency of the rest of the regions.

We further vary the LoRA rank ($r=\{4,8,16\}$) to analyze the trade-off between efficiency and representation capacity.  As shown in Table~\ref{tab:ablation}, increasing the rank consistently improves all metrics, with higher ranks yielding better background consistency and lower reconstruction errors.  Nevertheless, the model already performs competitively with a low rank ($r=4$), demonstrating that RegionRoute can achieve strong adaptation and generalization under highly compact low-rank constraints. This highlights that our design efficiently captures task-relevant subspaces while maintaining excellent parameter efficiency.

Overall, these results verify that each proposed component is crucial for fine-grained region control, and the complete RegionRoute configuration achieves the best balance between performance and parameter efficiency.
\section{Conclusion}
\label{sec:conclusion}

In this work, we presented an attention-supervised diffusion framework for precise and mask-free localized style transfer. By explicitly aligning the attention maps of style tokens with object masks during training, our method learns spatially grounded style associations without relying on handcrafted segmentation at inference. The proposed LoRA-MoE adaptation further enhances parameter efficiency and stylistic diversity, while our newly designed evaluation metric offers a more objective measure of localized style fidelity and identity preservation. Extensive experiments across multiple datasets demonstrate that our approach achieves controllable, semantically consistent, and high-quality regional stylization, advancing the practicality of diffusion-based visual editing. While our approach achieves promising results, it also opens several avenues for further exploration. Challenging cases such as very small, occluded, or semantically ambiguous objects reveal opportunities to strengthen attention-based spatial alignment. Additionally, extending the current framework beyond text-defined styles toward understanding and transferring styles from example images represents an exciting direction for future research.

{
    \small
    \bibliographystyle{ieeenat_fullname}
    \bibliography{main}
}

\clearpage
\setcounter{page}{1}
\maketitlesupplementary

\section{Pseudo Ground-Truth Generation}
We construct our training subset from the Grounded COCO dataset introduced in TokenCompose~\cite{tokencompose}, which augments MS-COCO~\cite{coco} image–caption pairs with object-level grounding. Concretely, we randomly sample 150 image–caption pairs from the training split to serve as our base data for LoRA-MoE fine-tuning~\cite{loramoe} and analysis.

For each selected image, we first choose a single target object to undergo localized style transfer. The target object is sampled uniformly at random from the annotated instances in that image. We then extract the corresponding binary segmentation mask at the original image resolution. Given the target image and its mask, we synthesize a pseudo ground-truth (pseudo-GT) image in a desired artistic style using a diffusion-based image-to-image style transfer model. Specifically, we employ Flux.1-Kontext~\cite{flux-kontext}: the original COCO image is used as the visual input, and the model is driven by a style instruction in the prompt, such as \textit{“make the image into pixel-art style”}, \textit{“make the image into cyberpunk style.”}. The diffusion model processes the entire image, producing a fully stylized version that preserves the global scene layout and object semantics while changing the overall appearance according to the requested style. After obtaining this stylized image, we use the target object mask to extract the corresponding stylized region and composite it back onto the original image. Concretely, we crop the stylized image using the binary mask, and replace the masked area in the original image with the stylized content.

Because no existing dataset provides direct supervision for localized and mask-conditioned style transfer, our pseudo-GT construction offers a practical way to obtain spatially grounded training pairs. Although the composited pseudo-GT images may contain imperfect boundaries or slight inconsistencies due to mask inaccuracies, this does not hinder the learning process. The Flux.1-Kontext~\cite{flux-kontext} model used in our fine-tuning exhibits strong semantic and object-level understanding, enabling it to correctly recognize and localize the target object even when the pseudo-GT supervision is not perfectly aligned. As a result, the model learns to generate smooth and coherent object boundaries during training and maintains robust object awareness at inference time, despite the approximate nature of the pseudo-GT masks.

\begin{figure*}[t]
    \centering
    \includegraphics[width=\linewidth]{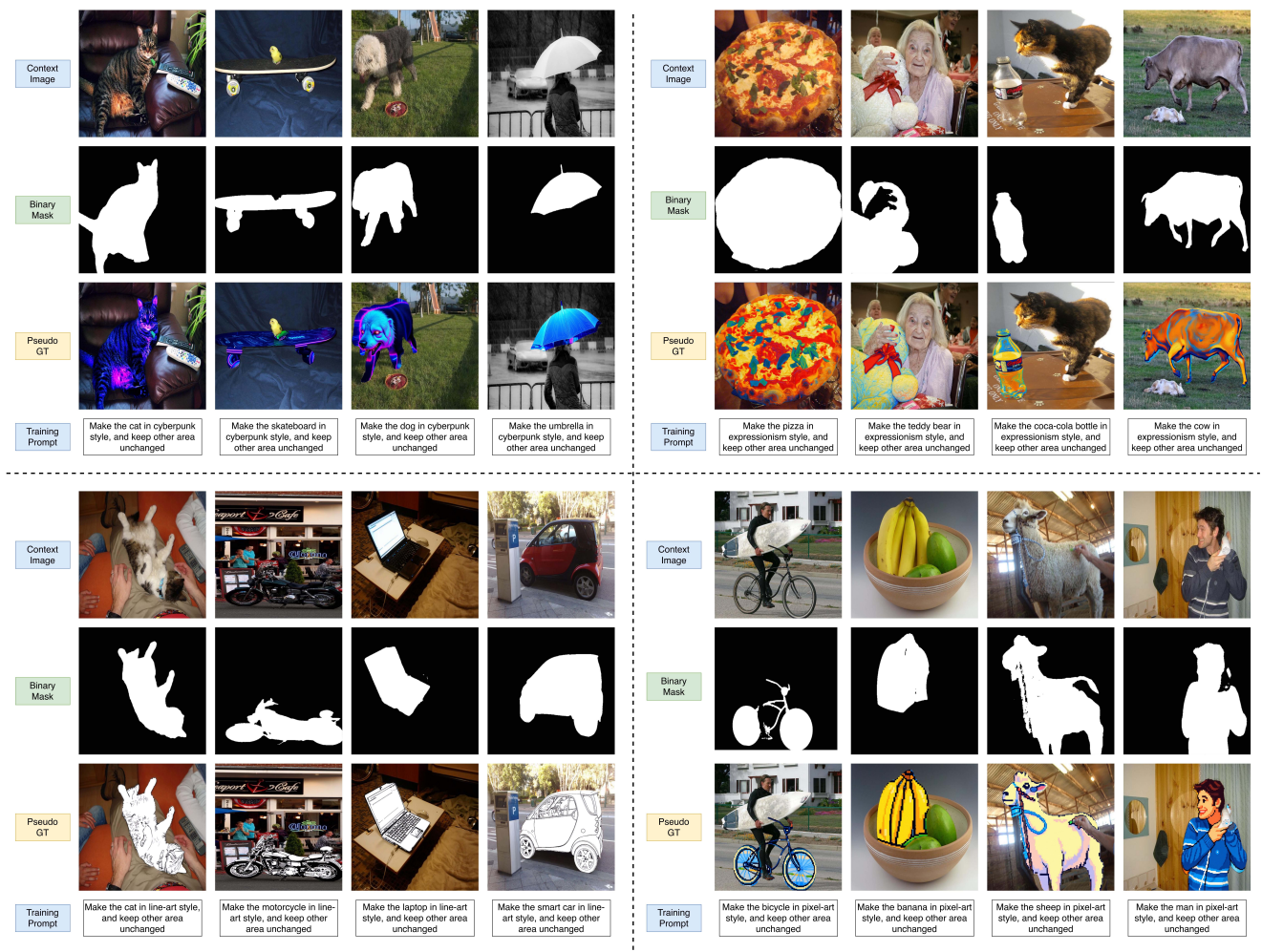}
    \caption{Examples from our pseudo ground truth dataset for localized style transfer. Each example consists of four elements: the original context image, the corresponding binary mask that specifies the target object, the pseudo ground truth image generated by applying a global style transformation to the entire image and compositing the stylized region back onto the original image, and the training prompt used for fine-tuning.}
    \label{fig:dataset}
\end{figure*}

To encourage stylistic diversity and avoid overfitting to a single visual domain, we generate pseudo-GT images in four representative styles: pixel art, cyberpunk, expressionism, and line art. For each of the 150 base images, we repeat the above procedure once for each style, using a style-specific instruction prompt (e.g., \textit{“make the image into line-art style”}) while keeping the underlying image and target object fixed. This yields four distinct pseudo-GT variants per image, resulting in a total of 600 stylized training samples (150 images × 4 styles). Each training sample thus consists of: (i) the original image, (ii) a binary mask for the target object, (iii) the corresponding pseudo-GT image where only the target region has been stylized, and (iv) a regional style editing instruction. Some training samples are shown in Figure~\ref{fig:dataset}. This construction provides the spatially localized supervision necessary for learning style transfer that is both content-aware and region-specific.

\begin{figure}[t]
    \centering
    \includegraphics[width=\linewidth]{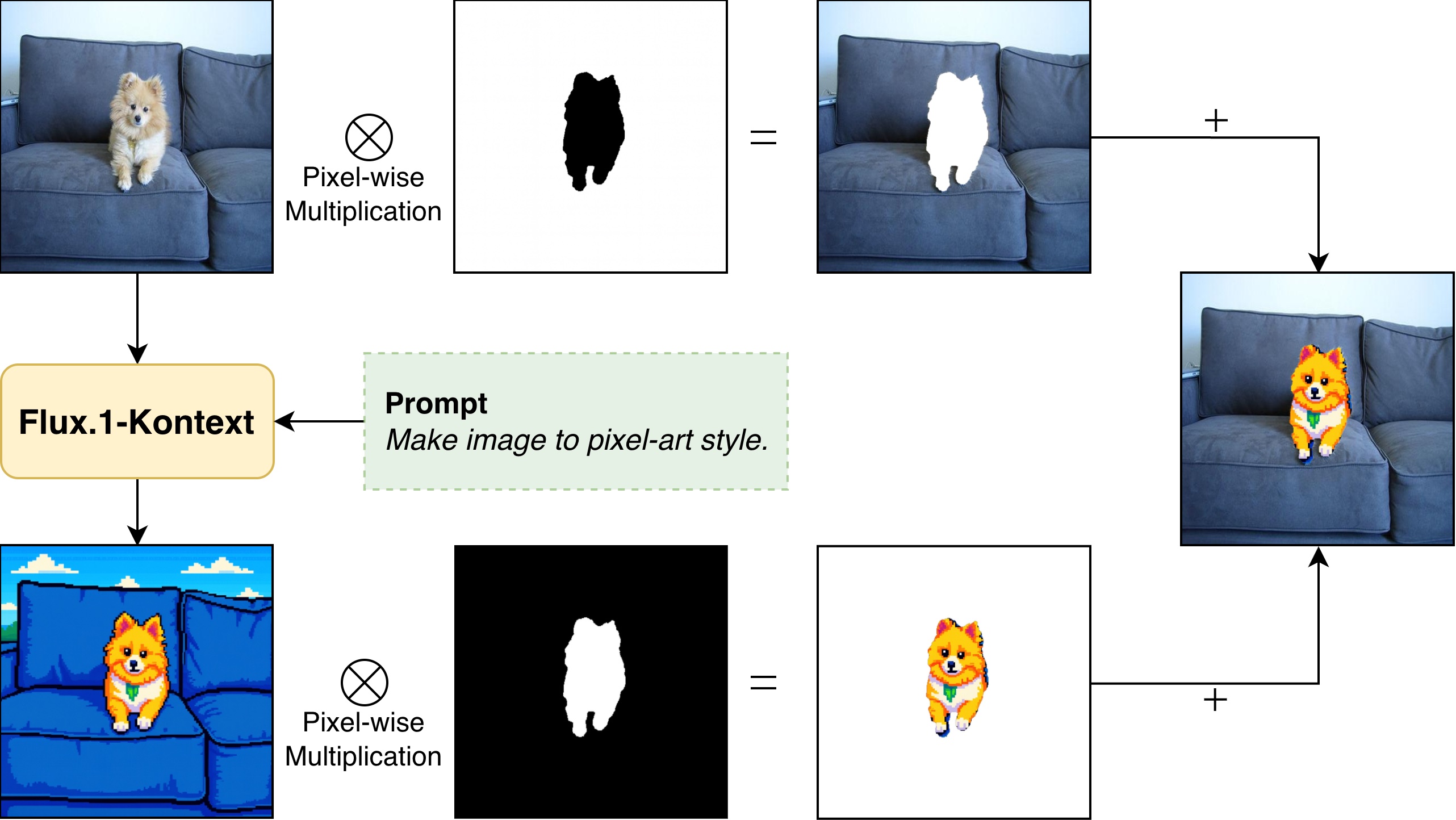}
    \caption{Illustration of the pseudo ground-truth (pseudo-GT) generation process. A diffusion-based style transfer model generates a fully stylized version of the image according to a given style prompt. The stylized region corresponding to the mask is then blended back into the original image using seamless cloning, producing an aligned input–target pair for localized style learning.}
    \label{fig:composition_pipeline}
\end{figure}

\begin{figure*}[t]
    \centering
    \includegraphics[width=0.95\linewidth]{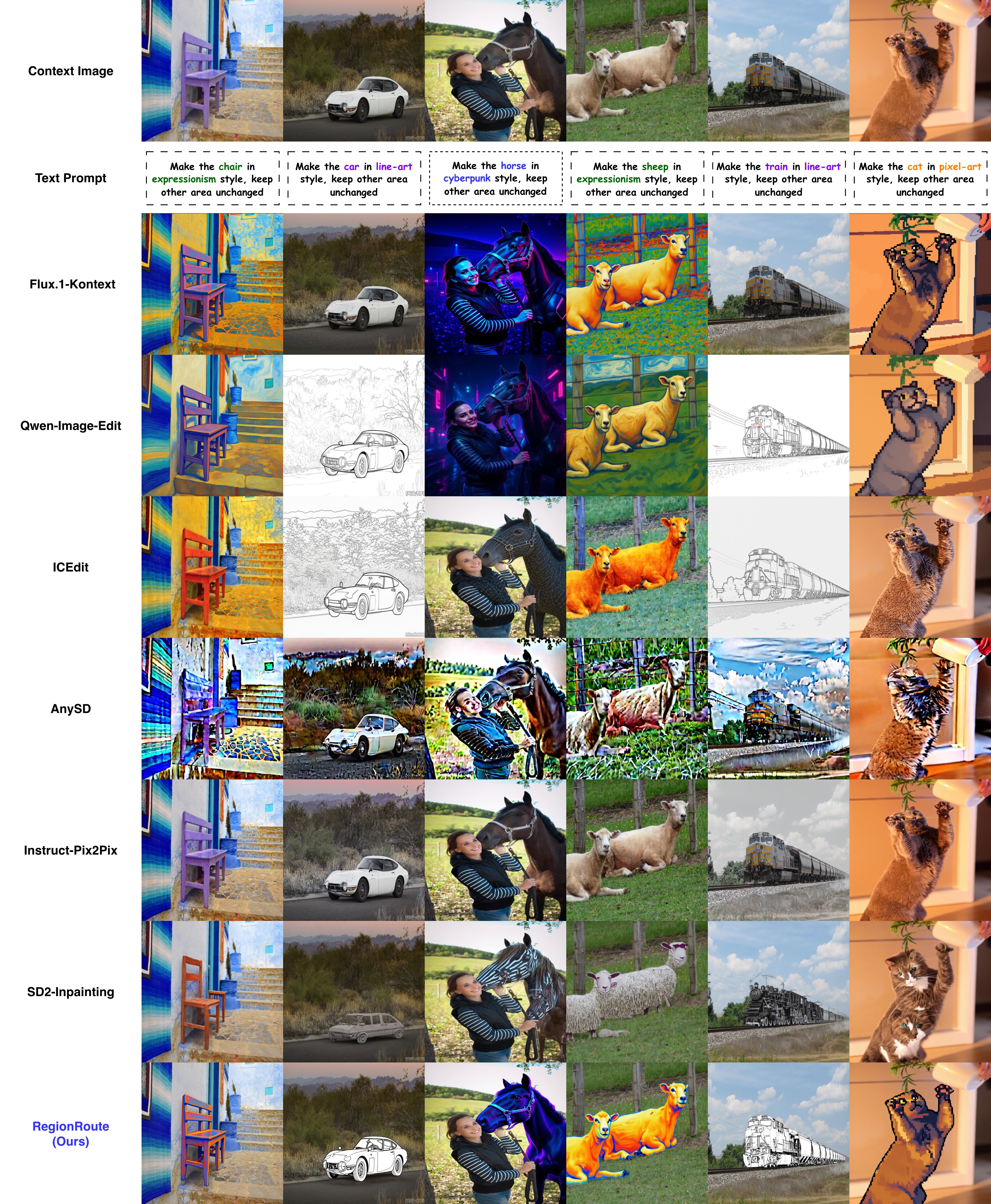}
    \caption{Additional examples of localized style transfer on objects that also appear in the training data. The model produces clean, structurally consistent stylization within the target regions, showing stable behavior on in-distribution categories.}
    \label{fig:sup_example1}
\end{figure*}

\section{More Examples and Analysis}

\begin{figure*}[t]
    \centering
    \includegraphics[width=0.95\linewidth]{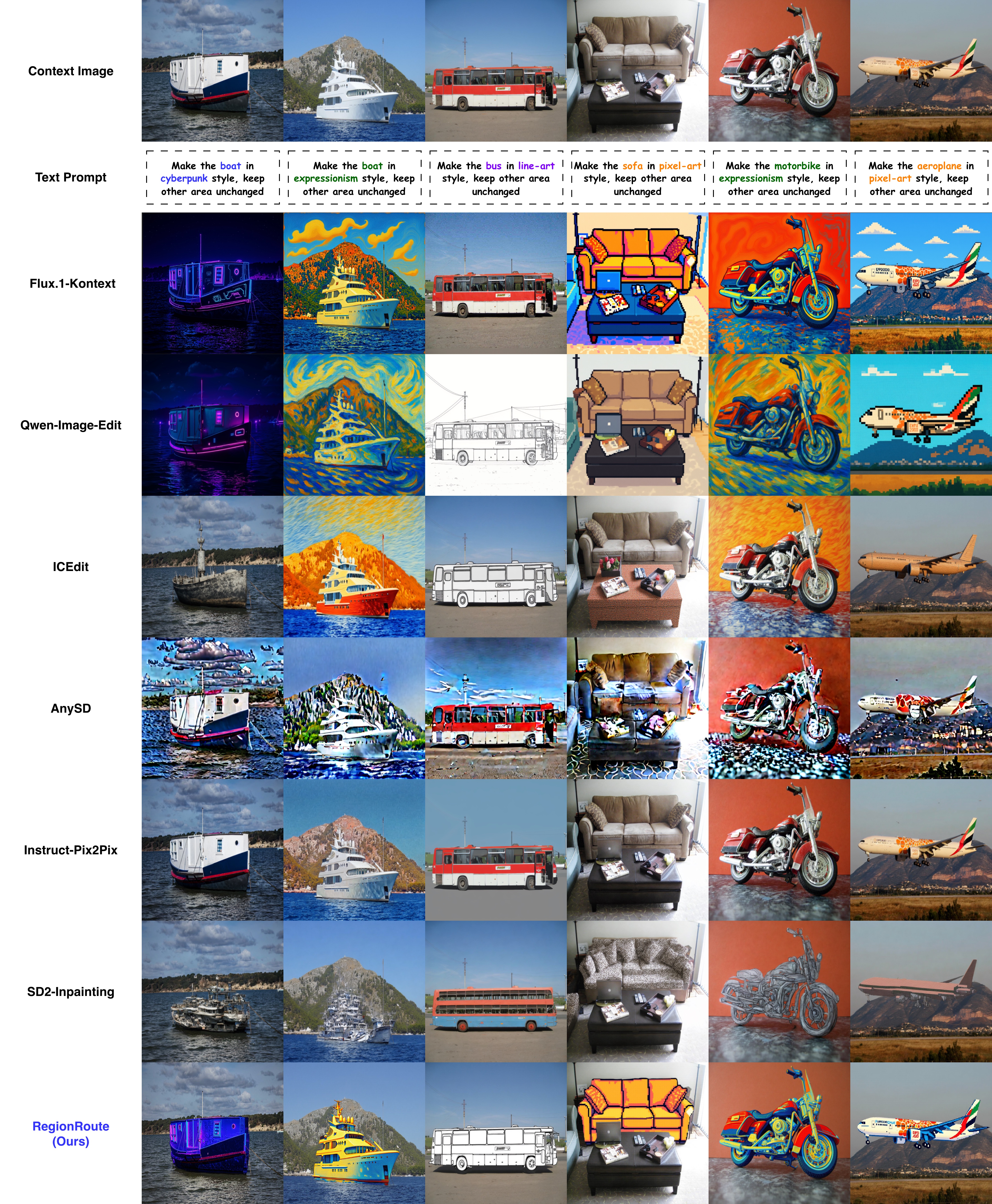}
    \caption{The first four columns show results on object categories that never appear in the training set. The last two show categories with only synonym-level presence (motorbike → motorcycle, aeroplane → airplane). The model successfully performs localized style transfer even without direct supervision for these object types, indicating strong semantic generalization.}
    \label{fig:sup_example2}
\end{figure*}

\subsection{Qualitative Analysis of Localized Style Transfer}
Figure~\ref{fig:sup_example1} shows additional qualitative results on object categories that appear in our training set. These examples correspond to the standard in-distribution case. As can be observed, the model consistently (i) identifies the target object region, (ii) applies the desired style within the masked area, and (iii) preserves the structural and geometric attributes of the object. The stylized region aligns well with the original content, and transitions at the mask boundaries remain visually smooth. These results further confirm that the model behaves reliably on object categories it has been exposed to during pseudo-GT fine-tuning.

Figure~\ref{fig:sup_example2} presents results on object categories that do not appear in the training set, providing additional evidence on the model’s generalization behavior. The first four examples involve categories entirely absent from training. The last two examples involve categories where only a synonym-level variant appears in training (e.g., motorbike appears only as motorcycle, and aeroplane appears only as airplane).

Despite the lack of category-level supervision during training, the model is able to accurately locate the target object and apply the style transformation in a structurally coherent manner. The transferred style remains consistent with the request while maintaining correct object semantics and boundaries. These results suggest that the model is not tied to specific object classes but instead has learned a more general mechanism for localized style transfer. The strong semantic understanding of the underlying Flux.1-Kontext~\cite{flux-kontext} model helps ensure that even imperfect pseudo-GT supervision does not significantly impair the model’s ability to generalize to new object categories.

\subsection{Qualitative Analysis for Ablation Study}

\begin{figure*}[t]
    \centering
    \includegraphics[width=\linewidth]{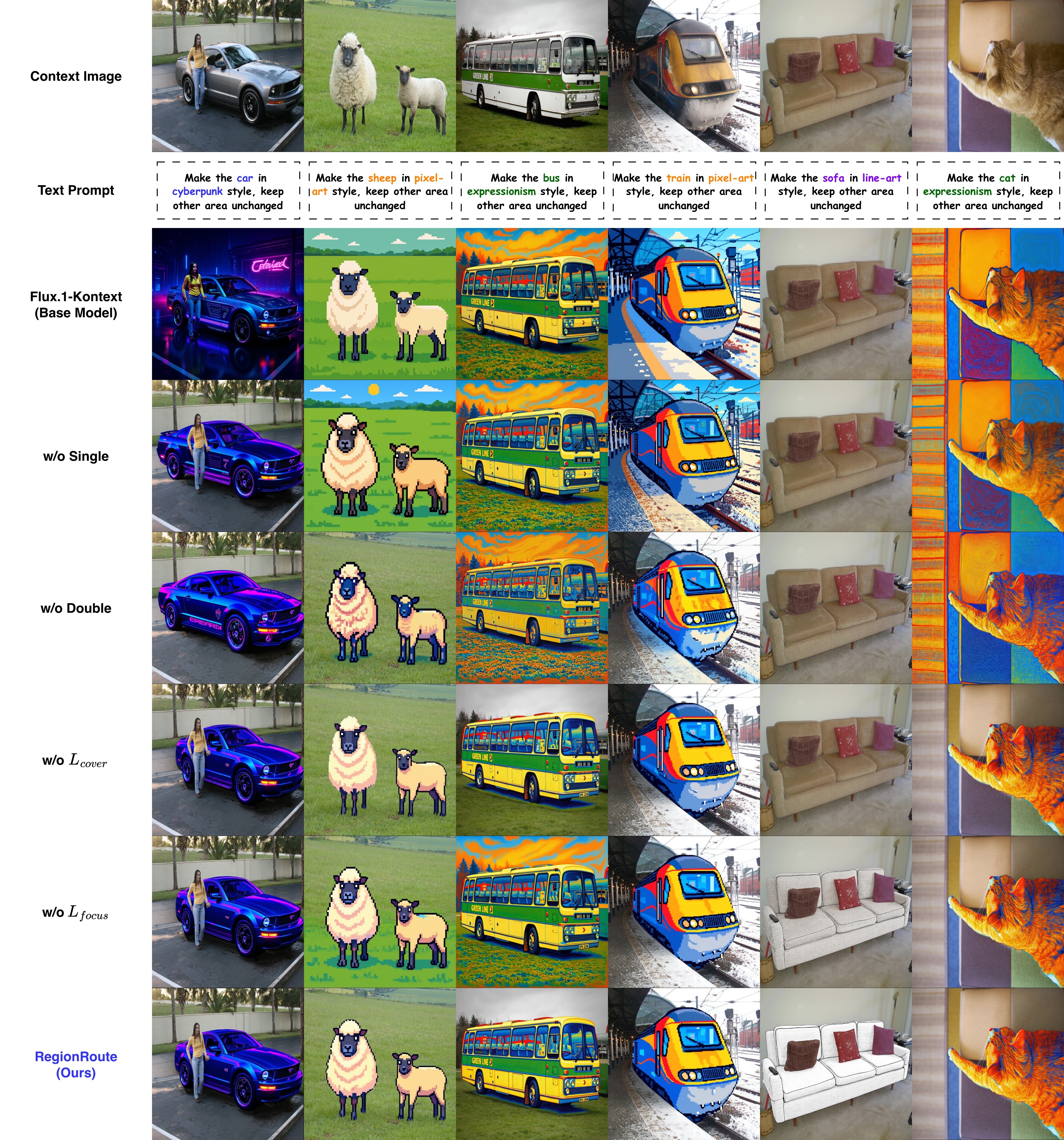}
    \caption{We evaluate the impact of removing single-stream LoRA blocks, double-stream LoRA blocks, the cover loss $\mathcal{L}_{\text{cover}}$, and the focus loss $\mathcal{L}_{\text{focus}}$. The full RegionRoute model achieves the clearest and most accurate localized style transfer, while ablated variants exhibit style leakage, incomplete stylization, or degraded unedited object consistency.}
    \label{fig:sup_example3}
\end{figure*}

Figure~\ref{fig:sup_example3} provides additional qualitative results comparing several ablated variants of RegionRoute. These examples help illustrate how different components contribute to regional style editing. As a reference, the original Flux.1-Kontext~\cite{flux-kontext} model is included. Without our designed modules, the base model tends to apply the requested style globally. While the stylization remains visually consistent, the absence of spatial selectivity highlights the need for additional mechanisms to enable region-level control.

When LoRA weights are removed from either the double stream blocks or the single stream blocks, the model is still able to stylize the target object in a number of cases, indicating that some degree of regional control is retained even when one LoRA branch is removed. However, these variants more frequently exhibit style leakage into non-target areas and less stable boundary behavior. The fact that both successful and unsuccessful examples appear suggests that the single stream and double stream LoRA fine-tuning each make a meaningful contribution to suppressing global style propagation during diffusion, with both branches providing complementary support for achieving cleaner regional editing.

We study the effect of removing the two training objectives used in our fine-tuning process. When the cover loss $\mathcal{L}_{\text{cover}}$ is removed, the model can still perform regional editing in a number of cases, but incomplete stylization becomes more common and parts of the target object may retain their original appearance. This indicates that $\mathcal{L}_{\text{cover}}$ encourages more uniform coverage within the masked region, although certain examples remain successful even without this objective. When the focus loss $\mathcal{L}_{\text{focus}}$ is removed, the model also retains the ability to produce correct localized stylization in several cases, but it shows a higher tendency to introduce spillover into the background. These behaviors suggest that both losses contribute to improving the consistency and spatial precision of regional editing, with each offering complementary benefits while not being strictly required for the model to succeed in some cases.

Overall, the full RegionRoute model, which combines both routing streams and both training objectives, produces the most consistent results across all examples. However, the ablations collectively show that each module contributes to regional style editing to some extent, and each variant retains certain successful cases. The observations suggest that style localization emerges from the combined effects of all components, with each module providing complementary improvements rather than acting as an isolated requirement.

\subsection{Qualitative Analysis for Attention Control}
\begin{figure*}[t]
    \centering
    \includegraphics[width=0.95\linewidth]{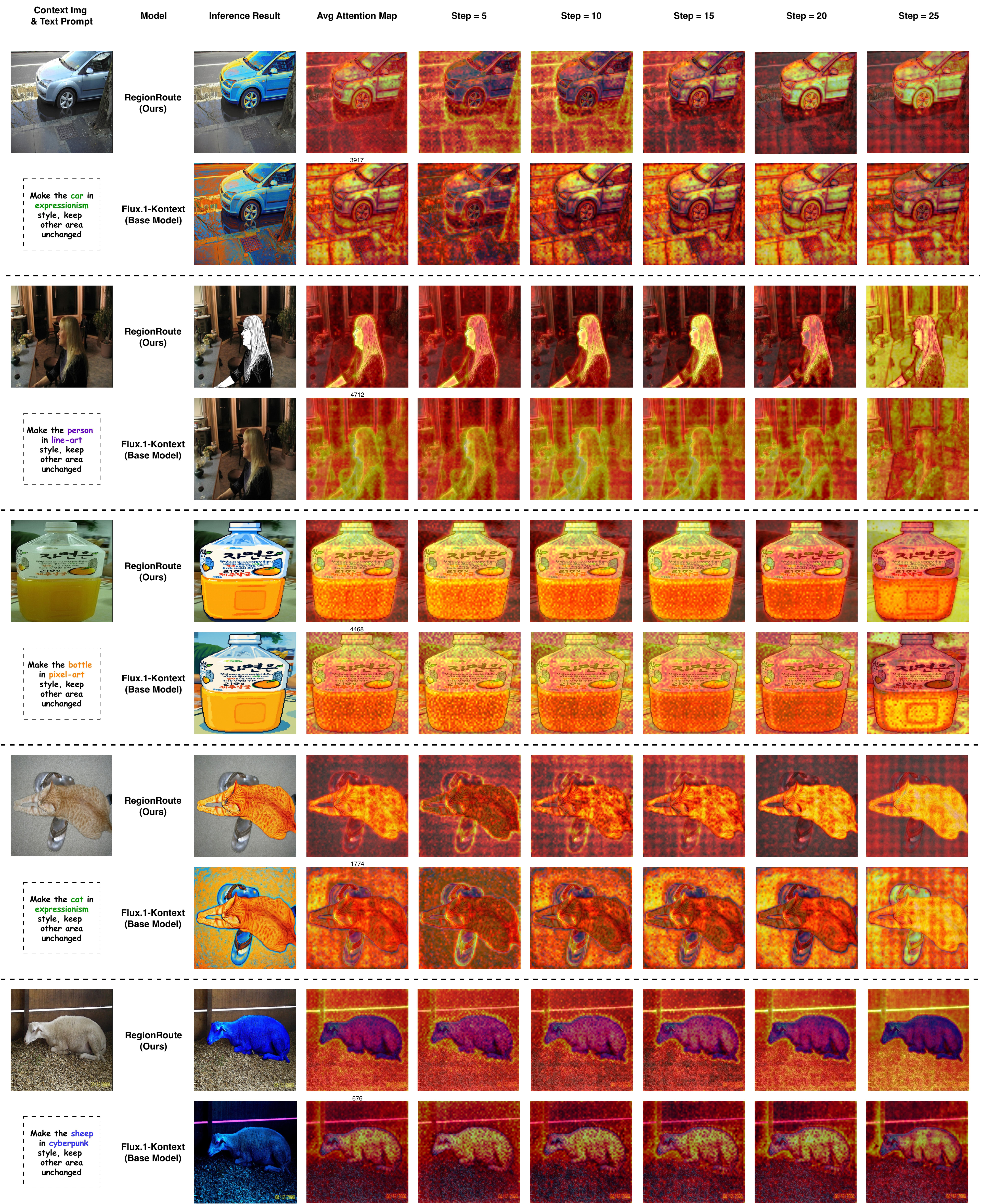}
    \caption{Comparison of the style-token attention maps for RegionRoute (ours) and the Flux.1-Kontext baseline. For each task, we show the edited output, the averaged style-token attention map, and the step-wise evolution of style-token attention at diffusion steps 5, 10, 15, 20, and 25. }
    \label{fig:sup_attn}
\end{figure*}

Figure~\ref{fig:sup_attn} analyzes how RegionRoute and the Flux.1-Kontext~\cite{flux-kontext} baseline allocate style-token attention during localized style-editing tasks. All visualizations correspond specifically to the attention from the style token toward spatial features at various diffusion steps, thereby revealing how each model propagates style information through the image.

Across all examples, RegionRoute demonstrates precise and stable routing of the style token to the intended object, producing well-defined attention patterns that match the object boundaries. In contrast, the baseline model consistently fails to form such a connection. Its style-token attention is diffuse, unstable, and often allocated to background regions or unrelated objects. This indicates that the baseline model lacks the capability to effectively anchor or attach the style token to the designated object, causing the style influence to spread globally rather than locally. As a result, the baseline’s final outputs exhibit undesirable global stylization and style leakage into regions that should remain unchanged.

The temporal attention evolution further reinforces this observation: RegionRoute maintains object-aligned style-token attention from early to late diffusion steps, whereas the baseline’s attention drifts or expands over time. These findings collectively demonstrate that RegionRoute provides a significant advantage in controlling the spatial propagation of style information by ensuring a persistent and accurate linkage between the style token and the target object.

\subsection{Qualitative Analysis for Failure Cases}

\begin{figure*}[t]
    \centering
    \includegraphics[width=\linewidth]{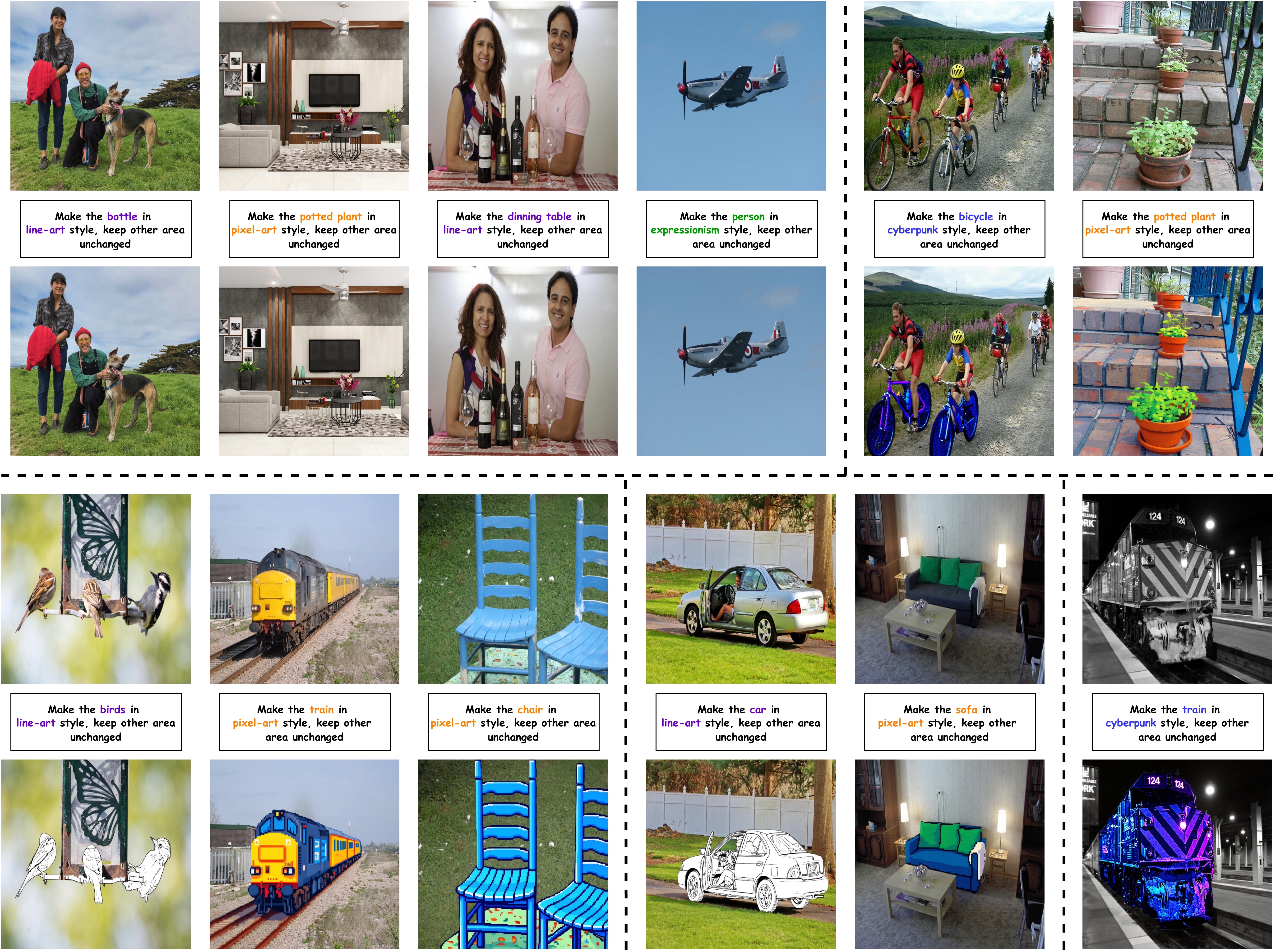}
    \caption{Examples illustrating several types of failure: (1) incomplete stylization when multiple target objects are present, where only the most salient instances are stylized; (2) failure to stylize very small or heavily occluded objects; (3) unintended stylization of regions adjacent to the target object, especially when objects are physically connected; (4) stylization leakage onto objects that are contained within the target object, such as a person inside a car or pillows on a sofa; and (5) partial stylization of a single object in cases where some object parts are difficult to recognize.}
    \label{fig:failure}
\end{figure*}

We summarize the major failure modes observed in our regional style transfer results, organized according to the underlying challenges. Figure~\ref{fig:failure} presents representative failure cases observed in our regional style transfer results. The first category involves small or hard-to-recognize objects. When the target object occupies very few pixels or is heavily occluded, such as a tiny potted plant, a small bottle, a distant person, or a table that is largely hidden behind other items, the model sometimes fails to identify the object, and the stylization does not occur. A related challenge arises when multiple instances of the target category appear in the same context image. In such cases, the model typically stylizes only the largest or most visually salient instances while neglecting smaller ones, suggesting that object-level recognition becomes unreliable for low-scale or low-visibility instances.

A second category of failures concerns unintended stylization of regions near the target object. When the target object is physically connected to or closely surrounded by other structures, the model may extend the stylization to these adjacent areas. Examples include stylizing portions of a bird perch along with the bird, a section of railway track along with a train, or the carpet underneath a chair. These behaviors indicate that spatial proximity and shared edges can sometimes be interpreted as part of a single coherent object.

A third type of failure involves objects that contain or enclose other objects, leading to ambiguous semantic grouping. When stylizing the car in an image where a person is sitting inside it, the person may also be stylized. Similarly, stylizing a sofa can unintentionally alter the appearance of pillows placed on it. These cases reflect ambiguity in the semantic interpretation of “the object” as described by the prompt.

Finally, we observe cases where only part of the target object is stylized. This typically occurs when certain regions of the object are visually ambiguous or difficult to separate from the background, resulting in incomplete localization.

Together, these failure modes illustrate the limitations that arise from object detection difficulty, small object scale, occlusion, close physical coupling of objects, and ambiguous semantic boundaries implied by user prompts.

\section{Details for VLM Evaluation}

We evaluate controllability and semantic reliability using the Qwen2.5-VL-7B-Instruct~\cite{qwen25} vision-language model (VLM). For each edited image, the VLM answers four binary questions. These questions assess whether the target style is correctly applied to the intended object, whether the background remains unchanged, and whether the model inadvertently introduces stylistic attributes unrelated to the user instruction.

For each sample, we load the target style and additionally sample a \emph{negative style}, which is randomly chosen from all styles except the target. The VLM then receives the edited image and answers:

\begin{itemize}
    \item \textbf{Q1:} ``Is the object in the \textit{target} style?''
    \item \textbf{Q2:} ``Is the background in the \textit{target} style?''
    \item \textbf{Q3:} ``Is the object in the \textit{negative} style?''
    \item \textbf{Q4:} ``Is the background in the \textit{negative} style?''
\end{itemize}

The negative style used in Q3--Q4 is deliberately unrelated to the target style. These two questions operate as sanity checks for both the editing model and the VLM. If the VLM tended to respond ``yes'' for arbitrary style queries, Q3--Q4 would expose such failure by producing uniformly high scores. In practice, Q3 and Q4 yield consistently low probabilities across almost all methods, demonstrating that the VLM is not biased toward affirmative answers and is capable of distinguishing stylistic attributes reliably.

For each image, the script constructs the four questions, packages them with the image into the VLM input, and enforces strict binary outputs by instructing the model to answer only ``yes'' or ``no.'' Responses are parsed and saved for aggregation, forming the probabilities reported in Table 2. This evaluation provides a scalable measure of object-level stylization accuracy, and background preservation.

\end{document}